\documentclass[10pt,twocolumn,letterpaper]{article}

\renewcommand\footnotemark{}

\usepackage{cvpr}
\usepackage{times}
\usepackage{epsfig}
\usepackage{graphicx}
\usepackage{amsmath}
\usepackage{amssymb}
\usepackage{color}
\usepackage{subfigure}
\usepackage{tabularx}
\usepackage{multirow}

\newcommand{\figref}[1]{Fig. \ref{#1}}
\newcommand{\tabref}[1]{Table \ref{#1}}
\newcommand{\equref}[1]{(\ref{#1})}
\newcommand{\secref}[1]{Sec. \ref{#1}}

\makeatletter
\def\hlinewd#1{%
	\noalign{\ifnum0=`}\fi\hrule \@height #1 \futurelet
	\reserved@a\@xhline}

\usepackage[breaklinks=true,bookmarks=false]{hyperref}

\cvprfinalcopy % *** Uncomment this line for the final submission

\def\cvprPaperID{6564} % *** Enter the CVPR Paper ID here

\begin{document}

%%%%%%%%% TITLE
\title{Semantic Attribute Matching Networks\thanks{This research was supported by R\&D program for Advanced Integrated-intelligence for Identification (AIID) through the National Research Foundation of KOREA (NRF) funded by Ministry of Science and ICT (NRF-2018M3E3A1057289).}}

\author{
	Seungryong Kim$^{1,2}$, Dongbo Min$^3$,  Somi Jeong$^1$, Sunok Kim$^{1,2}$, Sangryul Jeon$^1$, Kwanghoon Sohn$^{1,*}$\thanks{$^{*}$Corresponding author}\\
	$^1$Yonsei University, 
	$^2$École Polytechnique Fédérale de Lausanne (EPFL), 
	$^3$Ewha Womans University\\
	\texttt{seungryong.kim@epfl.ch}\\
	\texttt{\{somijeong,kso428,cheonjsr,khsohn\}@yonsei.ac.kr},
	\texttt{dbmin@ewha.ac.kr}\\
} 

\maketitle
%\thispagestyle{empty}

%%%%%%%%% ABSTRACT
\begin{abstract}
We present semantic attribute matching networks (SAM-Net) for jointly establishing correspondences and transferring attributes across semantically similar images, which intelligently
weaves the advantages of the two tasks while overcoming their limitations. SAM-Net accomplishes this through an iterative process of establishing reliable correspondences by reducing
the attribute discrepancy between the images and synthesizing attribute transferred images using the learned correspondences. 
To learn the networks using weak supervisions in the form of image pairs, we present a semantic attribute matching loss based on the matching similarity between
an attribute transferred source feature and a
warped target feature. With SAM-Net, the state-of-the-art performance is attained on several benchmarks for semantic matching and attribute transfer.
\end{abstract}

\section{Introduction}\label{sec:1}
\begin{figure}[t]
	\centering
	\renewcommand{\thesubfigure}{}
	\subfigure[]
	{\includegraphics[width=1\linewidth]{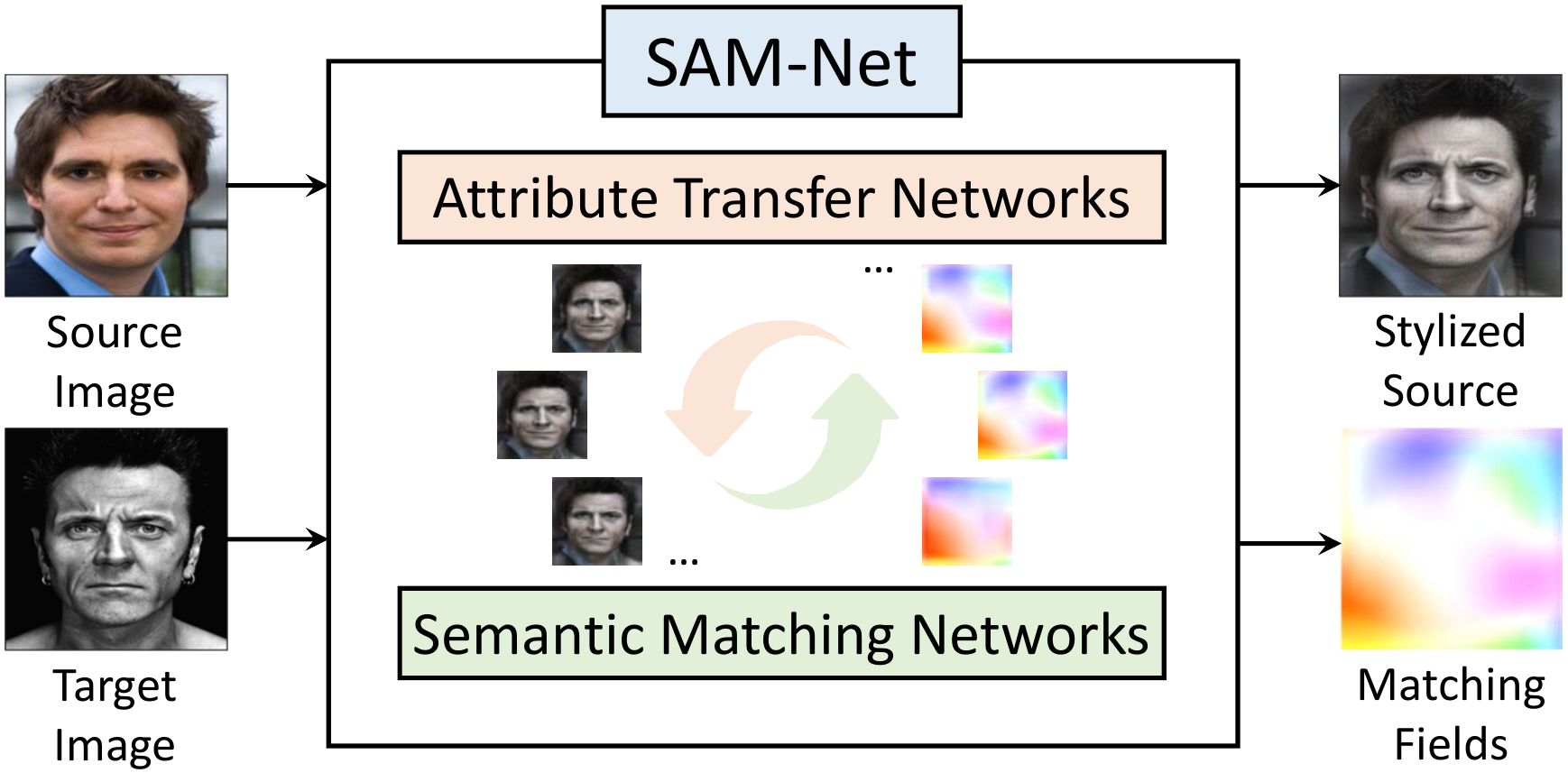}}\\
	\vspace{-10pt}	
	\caption{Illustration of SAM-Net: for semantically similar images having both photometric and geometric variations, SAM-Net recurrently estimates semantic correspondences and synthesizes attribute transferred images in a joint and boosting manner.}\label{img:1}\vspace{-10pt}
\end{figure}

Establishing correspondences and transferring attributes across \emph{semantically} similar images can facilitate a variety of computer vision applications~\cite{Liu11,Liao17,Kim18}. 
In these tasks, the images resemble each other in contents but differ in visual attributes, such as color, texture, and style, e.g., the images with different faces as exemplified in~\figref{img:1}. Numerous techniques have been proposed for the semantic correspondence~\cite{Han17,Kim17,Rocco18,Jeon18,Rocco18nips,Kim18nips} and attribute transfer~\cite{Gatys16,Chen16,Li16,Johnson16,Luan17,Huang17,Jing2017,Huang17,Liao17,Gu18}, but these two tasks have been studied independently although they can be mutually complementary.

To establish reliable semantic correspondences, state-of-the-art methods have leveraged deep convolutional neural networks (CNNs) in extracting descriptors~\cite{Choy16,Zhou16,Kim17} and regularizing correspondence fields~\cite{Han17,Rocco18,Jeon18,Rocco18nips,Kim18nips}. 
Compared to conventional handcrafted methods~\cite{Liu11,Kim13,Bristow15,Zhou15,Taniai16}, 
they have achieved a highly reliable performance. 
To overcome the problem of limited ground-truth supervisions, some methods~\cite{Rocco18,Jeon18,Rocco18nips,Kim18nips} have tried to learn deep networks using only weak supervision in the form of image pairs based on the intuition that the matching cost between the source and target features over a set of
transformations should be minimized at the correct transformation. These methods presume that the attribute
variations between source and target images are negligible
in the deep feature space. However, in practice the
deep features often show limited performance in handling
different attributes that exist in the source and target images,
often degrading the matching accuracy dramatically.

To transfer the attributes between source and target images, following the seminal work of Gatys et al.~\cite{Gatys15}, numerous methods have been proposed to separate and recombine the contents and attributes using deep CNNs~\cite{Gatys16,Chen16,Li16,Johnson16,Luan17,Huang17,Jing2017,Huang17,Liao17,Gu18}. 
%These methods can be generally classified into parametric and non-parametric approaches. 
Unlike the parametric methods~\cite{Gatys16,Johnson16,Luan17,Huang17} that match the global statistics of deep features while ignoring the spatial layout of contents, the non-parametric methods~\cite{Chen16,Li16,Liao17,Gu18} directly find neural patches in the target image similar to the source patch and synthesize them to reconstruct the stylized image. 
These non-parametric methods generally estimate nearest neighbor
patches between source and target images with weak implicit regularization methods~\cite{Chen16,Li16,Liao17,Gu18} using a simple local aggregation followed by a winner-takes-all (WTA). 
%These methods were mainly tailored to artistic style transfer~\cite{Gatys16,Chen16,Johnson16}. 
However, photorealistic attribute transfer needs highly regularized and semantically meaningful correspondences, and thus existing methods~\cite{Chen16,Li16,Gu18} frequently fail when the images have background clutters and different attributes while representing similar global feature statistics. A method called deep image analogy~\cite{Liao17} has tried to estimate more semantically meaningful dense corrrespondences for photorealistic attribute transfer, but it still has limited localization ability with PatchMatch~\cite{Barnes09}.

In this paper, we present semantic attribute matching networks (SAM-Net) for overcoming the aforementioned limitations of current semantic matching and attribute transfer techniques. 
The key idea is to weave the advantages of semantic matching and attribute transfer networks in a boosting manner. Our networks accomplish this through an iterative process of establishing more reliable semantic correspondences by reducing the attribute discrepancy between semantically similar images and synthesizing an attribute transferred image with the learned semantic correspondences. Moreover, our networks are learned from weak supervision in the form of image pairs using the proposed semantic attribute matching loss. Experimental results show that SAM-Net outperforms the latest methods for semantic matching and attribute transfer on several benchmarks, including TSS dataset~\cite{Taniai16}, PF-PASCAL dataset~\cite{Ham17}, and CUB-200-2011 dataset~\cite{WahCUB_200_2011}.

\section{Related Work}\label{sec:2}
\paragraph{Semantic correspondence.}
Most conventional methods for semantic correspondence that use handcrafted features and regularization methods~\cite{Liu11,Kim13,Bristow15,Zhou15,Taniai16} have provided limited performance due to a low discriminative power. Recent approaches have used deep CNNs for extracting their features~\cite{Choy16,Zhou16,Kim17,Novotny17} and regularizing correspondence fields~\cite{Han17,Rocco17,Rocco18}. 
%For instance, there are techniques that utilize a 3D CAD model for supervision~\cite{Zhou16}, employ fully convolutional feature learning~\cite{Choy16}, learn filters with geometrically consistent responses across different object instances~\cite{Novotny17}, exploit local self-similarity within a fully convolutional network~\cite{Kim17}, and estimate correspondences using object proposals~\cite{Ham16,Ham17,Han17,Ufer17}. 
Rocco et al.~\cite{Rocco17,Rocco18} proposed deep architecture for estimating a geometric matching model, but these methods estimate only globally-varying geometric fields.
%thus leading to limited performance in dealing with locally-varying geometric deformations. 
%None of these methods is able to handle non-rigid geometric variations. 
To deal with locally-varying geometric deformations, some methods such as UCN~\cite{Choy16} and CAT-FCSS~\cite{Kim18} were proposed based on STNs~\cite{Jaderberg15}. Recently, PARN~\cite{Jeon18}, NC-Net~\cite{Rocco18nips}, and RTNs~\cite{Kim18nips} were proposed to estimate locally-varying transformation fields using a coarse-to-fine scheme~\cite{Jeon18}, neighbourhood consensus~\cite{Rocco18nips}, and an iteration technique~\cite{Kim18nips}. These methods~\cite{Jeon18,Rocco18nips,Kim18nips} presume that the attribute variations between source and target images are negligible in the deep feature space. However, in practice the deep features often show limited performance in handling different attributes. Aberman et al.~\cite{Aberman18} presented a method to deal with the attribute variations between the images using a variant of instance normalization~\cite{Huang17}. However, the method does not have an explicit learnable module to reduce the attribute discrepancy, thus yielding limited performance.
\vspace{-10pt}
\begin{figure*}[t]
	\centering
	\renewcommand{\thesubfigure}{}
	\subfigure[(a)]
	{\includegraphics[width=0.31\linewidth]{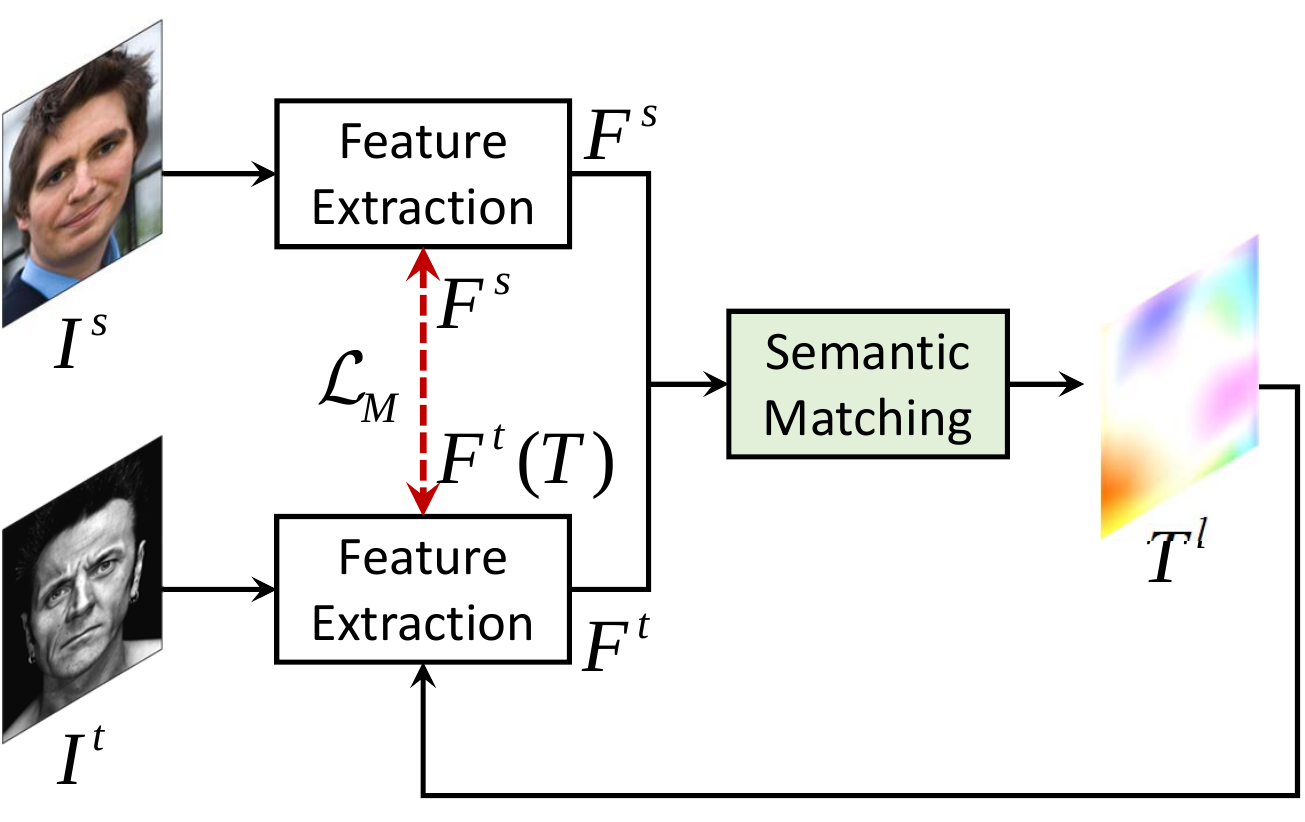}}\hfill
	\subfigure[(b)]
	{\includegraphics[width=0.31\linewidth]{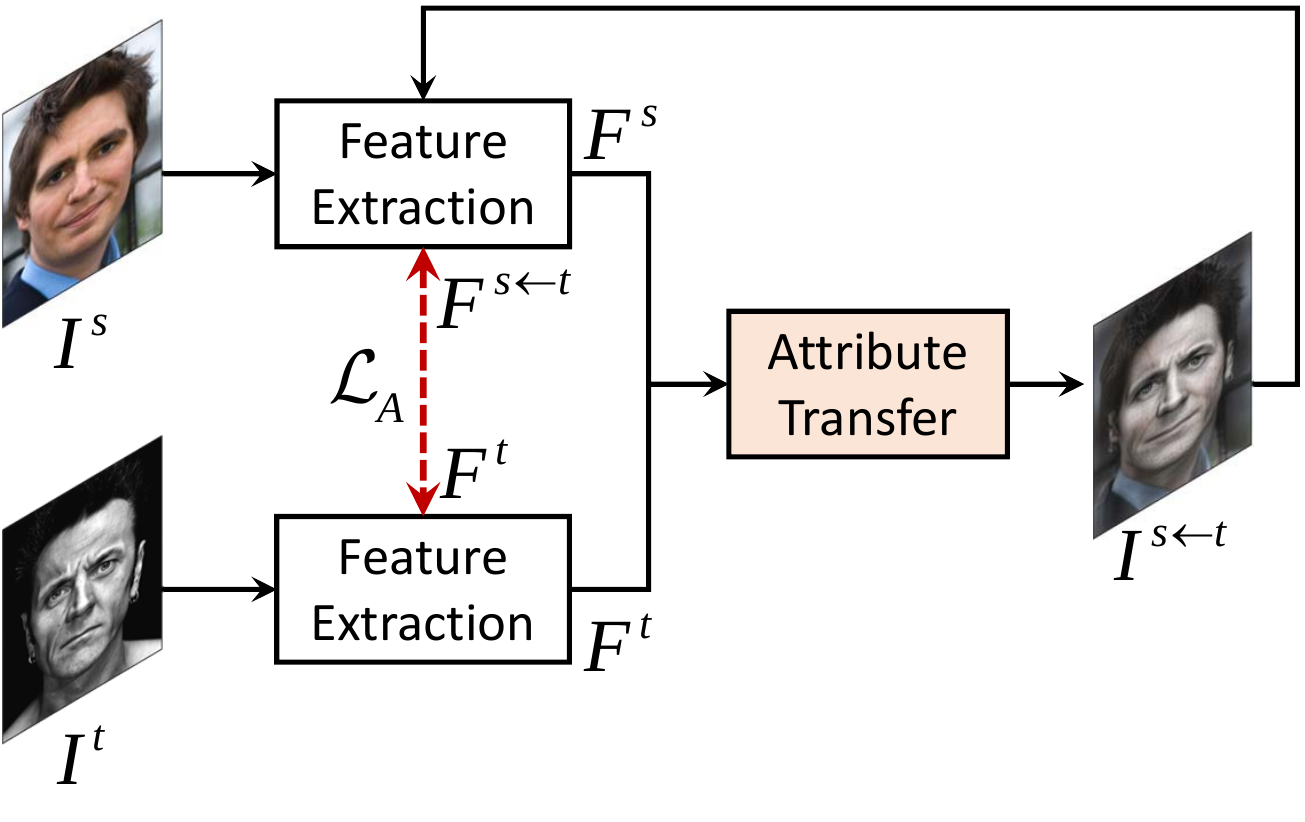}}\hfill
	\subfigure[(c)]
	{\includegraphics[width=0.31\linewidth]{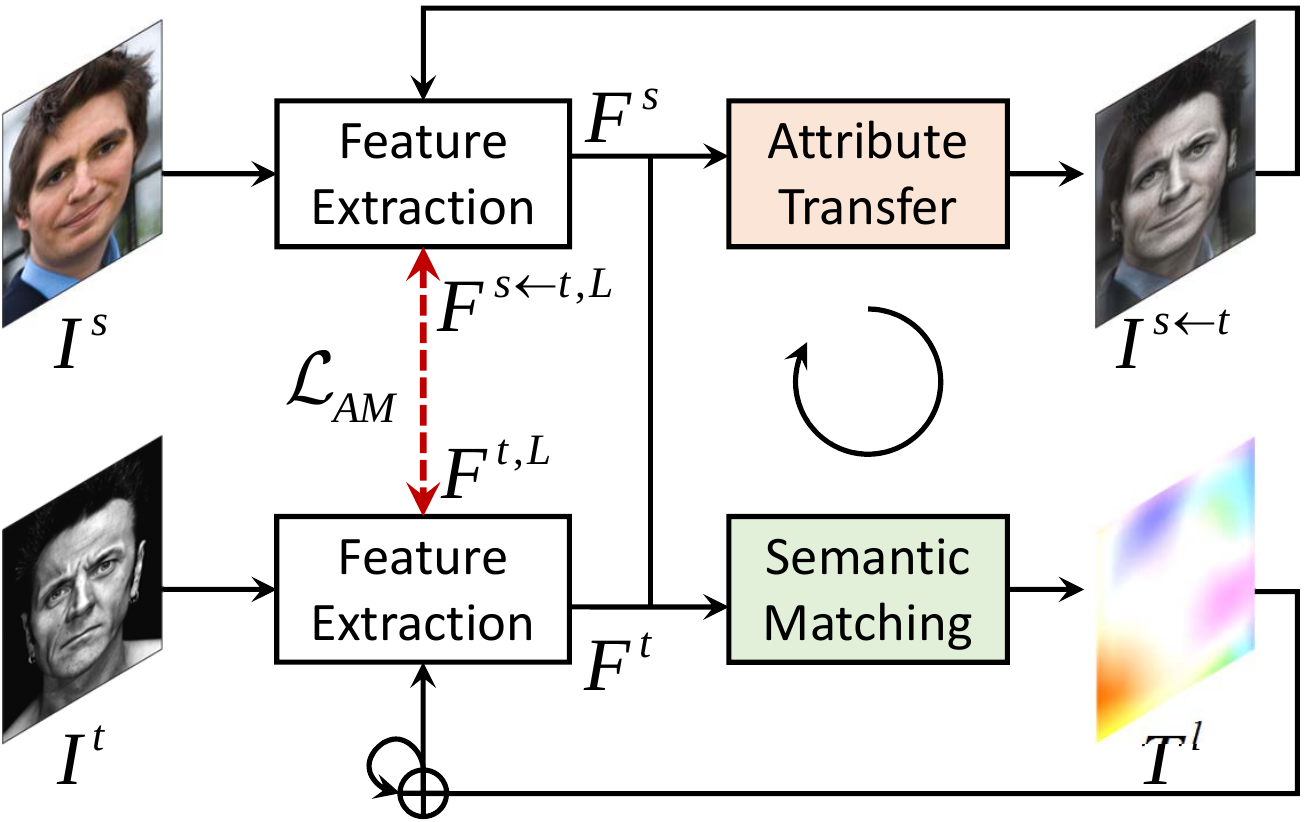}}\hfill\\
	\caption{Intuition of SAM-Net: (a) methods for semantic matching~\cite{Rocco17,Rocco18,Kim18nips,Jeon18}, (b) methods for attribute transfer~\cite{Gatys16,Johnson16,Li16}, and (c) SAM-Net, which recurrently weaves the advantages of both existing semantic matching and attribute transfer techniques.}\label{img:2}\vspace{-10pt}
\end{figure*} 

\paragraph{Attribute transfer.}
There have been a lot of works on the transfer of visual attributes, e.g., color, texture, and style, from one image to another, and most approaches are tailored to their specific objectives~\cite{Reinhard01,Tai05,Efros01,Ashikhmin03,Zhang13,Frigo16}. Since our method represents and synthesizes deep features to transfer the attribute between semantically similar images, the neural style transfer~\cite{Gatys16,Chen16,Johnson16,Jing2017} is highly related to ours. In general, these approaches can be classified into parametric and non-parametric methods. 

In parametric methods, inspired by the seminal work of Gatys et al.~\cite{Gatys15}, numerous methods have been presented, such as the work of Johnson et al.~\cite{Johnson16}, AdaIN~\cite{Huang17}, and WCT~\cite{Li17}. Since these methods are globally formulated, they have shown limited performance for photorealistic stylization tasks~\cite{Li18,Luan17}. To alleviate these limitations, Luan et al. proposed a deep photo style transfer~\cite{Luan17} that computes and uses the semantic labels. Li et al. proposed Photo-WCT~\cite{Li18} to eliminate the artifacts using additional smoothing step. However, these methods still have been formulated without considering semantically meaningful correspondence fields. 

Among non-parametric methods, the seminal work of Li et al.~\cite{Li16} first searches local neural patches, which are similar to the patch of content image, in the target style image to preserve the local structure prior of content image, and then uses them to synthesize the stylized image. Chen et al.~\cite{Chen16} sped up this process using the feed-forward networks to decode the synthesize features. Inspired by this, various approaches have been proposed to synthesize locally blended features efficiently~\cite{Li16b,Ulyanov16,Lu17,Li17b,Ulyanov17}. However, the aforementioned methods are tailored to the artistic style transfer, and thus they focused on finding the patches to reconstruct more plausible images, rather than finding semantically meaningful dense correspondences. They generally estimate the nearest neighbor patches using weak implicit regularization methods such as WTA. 
Recently, Gu et al.~\cite{Gu18} introduced a deep feature reshuffle technique to connect both parametric and non-parametric methods, but they search the nearest neighbor using an expectation-maximization (EM) that also produces limited localization accuracy.

More related to our work is a method called deep image analogy~\cite{Liao17} that searches semantic correspondences using deep PatchMatch~\cite{Barnes09} in a coarse-to-fine manner. However, PatchMatch inherently has a limited regularization power as shown in~\cite{Kim17c,Lu13,Li15}. In addition, the method still needs the greedy optimization for feature deconvolution that induces computational bottlenecks, and only considers the translational fields, thus having the limitation to handle more complicated deformations. 
%By contrast, our method formulates the semantic matching and photorealistic attribute transfer within deep CNNs in an end-to-end and iterative manner for jointly estimating the correspondences and synthesizing the attribute transferred images efficiently and effectively.

\section{Problem Statement}\label{sec:3}
Let us denote \emph{semantically} similar source and target images as $I^s$ and $I^t$, respectively. The objective of our method is to jointly establish a correspondence field $f_i=[u_i,v_i]^T$ between the two images that is defined for each pixel $i=[i_\mathrm{x},i_\mathrm{y}]^T$ and synthesize an attribute transferred image $I^{s\leftarrow t}$ by transferring an attribute of target image $I^t$ to a content of source image $I^s$. 
%Although the semantic correspondence and attribute transfer tasks can be mutually complementary, these two tasks have been independently studied.

CNN-based methods for semantic correspondence~\cite{Rocco17,Kim18,Rocco18,Jeon18,Rocco18nips,Kim18nips} involve first extracting deep features~\cite{Simonyan15,Kim18}, denoted by $F^s_i$ and $F^t_i$, from $I^s_i$ and $I^t_i$ within local receptive fields, and then estimating correspondence field $f_i$ of the source image using deep regularization models~\cite{Rocco17,Rocco18,Kim18nips}, as shown in~\figref{img:2}(a). To learn the networks using only image pairs, some methods~\cite{Rocco18,Kim18nips} formulate the loss function based on the intuition that the matching cost between the source feature $F^s_i$ and the target feature $F^t_{i+f_i}$ over a set of transformations should be minimized. For instance, they formulate the matching loss defined as
\begin{equation}
\mathcal{L}_{M} = \sum\nolimits_{i}{\|F^s_i-F^t_{i+f_i}\|^2_F},
\end{equation}
where $\|\cdot\|^2_F$ denotes Frobenius norm. To deal with more complex deformations such as affine transformation~\cite{Kim17c,Kim18nips}, instead of $F^t_{i+f_i}$, $F^t({T}_i)$ or $F^t_{i+f_i}(A_i)$ can be used with a $2\times 3$ matrix ${T}_i = [{A}_{i},{f}_i]$. 
Although semantically similar images can share similar contents but have different attributes, these methods~\cite{Rocco17,Rocco18,Jeon18,Rocco18nips,Kim18nips} simply assume that the attribute variations between source and target images are negligible in the deep feature space. It thus cannot guarantee measuring a fully accurate matching cost without an explicit module to reduce the attribute gaps.

To minimize the attribute discrepancy between source and target images, attribute or style transfer methods~\cite{Gatys16,Chen16,Johnson16,Jing2017} separate and recombine the content and attribute. Unlike the parametric methods~\cite{Gatys16,Luan17}, the non-parametric methods~\cite{Chen16,Li16,Liao17,Gu18} directly find neural patches in the target image similar to the source patch and synthesize them to reconstruct the stylized feature $F^{s\leftarrow t}$ and image $I^{s\leftarrow t}$, as shown in~\figref{img:2}(b). Formally, they formulate two loss functions including the content loss defined as
\begin{equation}\label{equ:loss_cont}
\mathcal{L}_{C} = \sum\nolimits_{i}{\|F^{s\leftarrow t}_i-F^s_i\|^2_F},
\end{equation}
and the non-parametric attribute transfer loss defined as
\begin{equation}
\mathcal{L}_{A}
= \sum\nolimits_{i}{\sum\nolimits_{j\in N_i}{\|F^{s\leftarrow t}_j-F^t_{j+f_i}\|^2_F}},
\end{equation}
where ${i+f_i}$ is the center point of the patch in $I^t$ that is most similar to a patch centered at $i$ in $I^s$. 
%Note that patch-level mapping is used, instead of pixel-level mapping, to transfer the local statistical characteristic more effectively~\cite{Chen16,Li16,Liao17,Gu18}. 
Generally, ${f_i}$ is determined using the matching scores of normalized cross-correlation~\cite{Chen16,Li16} aggregated on $N_i$ over all local patches followed by the labeling optimization such that 
\begin{equation}\label{equ:NCC}
f_i = 
\mathop{\mathrm{argmax}}_{m}{\sum\nolimits_{j\in N_i}{({F^s_j \boldsymbol{\cdot} F^t_{j+m}})/\|F^s_j\|\|F^t_{j+m}\| }},
\end{equation}
where the operator $\boldsymbol{\cdot}$ denotes inner product.

However, the hand-designed discrete labeling techniques such as WTA~\cite{Chen16,Li16}, PatchMatch~\cite{Liao17}, and EM~\cite{Gu18} used to optimize \equref{equ:NCC} rely on weak implicit smoothness constraints, often producing poor matching results. In addition, they only consider the translational fields, i.e., $f_i$, thus limiting handling more complicated deformations caused by scale, rotation and skew that may exist among object instances. 
%Using the global optimizers may mitigate such matching outliers in the semantic correspondence~\cite{Kim17c}, but their accuracy is generally inferior to recent deep semantic correspondence approaches~\cite{Rocco18,Kim18nips} that work end-to-end.
\begin{figure*}
	\centering
	\renewcommand{\thesubfigure}{}
	\subfigure[]
	{\includegraphics[width=1\linewidth]{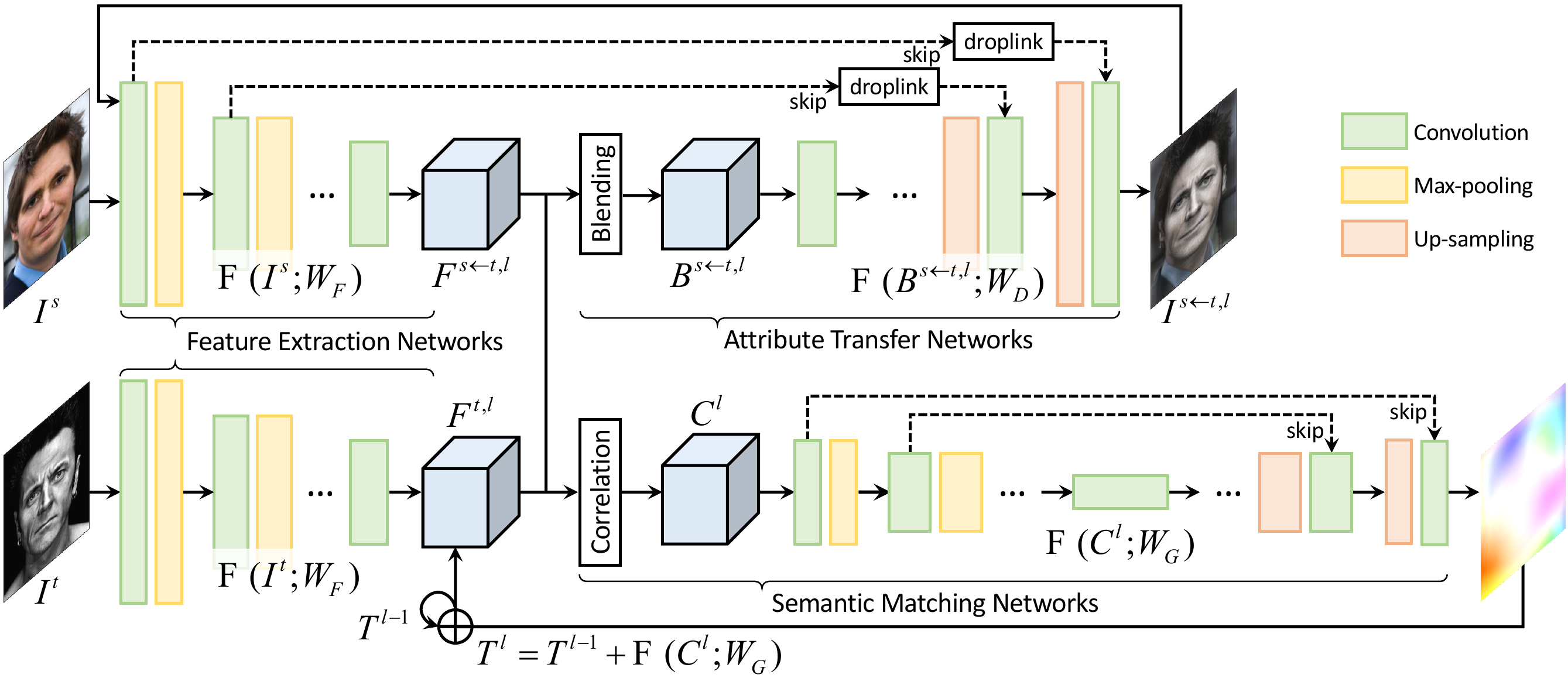}}\\
	\vspace{-10pt}
	\caption{Network configuration of SAM-Net, consisting of feature extraction networks, semantic matching networks, and attribute transfer networks in a recurrent structure. Initially, $F^{s\leftarrow t,0}=F^s$ and $F^{t,0}=[\mathrm{I}_{2\times2},\mathrm{0}_{2\times1}]$. They output $T^l_i$ and $I^{s\leftarrow t,l}$ at each $l$-th iteration. }\label{img:3}\vspace{-10pt}
\end{figure*}
\begin{figure*}[t]
	\centering
	\renewcommand{\thesubfigure}{}
	\subfigure[(a)]
	{\includegraphics[width=0.122\linewidth]{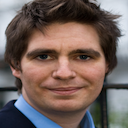}}\hfill
	\subfigure[(b)]
	{\includegraphics[width=0.122\linewidth]{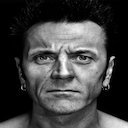}}\hfill
	\subfigure[(c)]
	{\includegraphics[width=0.122\linewidth]{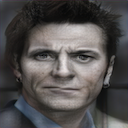}}\hfill
	\subfigure[(d)]
	{\includegraphics[width=0.122\linewidth]{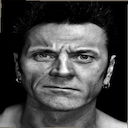}}\hfill
	\subfigure[(e)]
	{\includegraphics[width=0.122\linewidth]{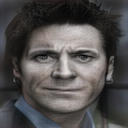}}\hfill
	\subfigure[(f)]
	{\includegraphics[width=0.122\linewidth]{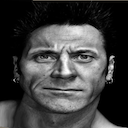}}\hfill
	\subfigure[(g)]
	{\includegraphics[width=0.122\linewidth]{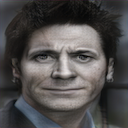}}\hfill
	\subfigure[(h)]
	{\includegraphics[width=0.122\linewidth]{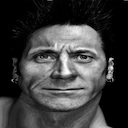}}\hfill
	\caption{Convergence of SAM-Net: (a) source image, (b) target image, iterative evolution of attribute transferred images (c), (e), and (g) and warped images using dense corresondences (d), (f), and (h) after iteration 1, 2, and 3. In the recurrent formulation of SAM-Net, the predicted transformation fields and attribute transferred images become progressively more accurate through iterative estimation.}\label{img:6}\vspace{-10pt}
\end{figure*}

\section{Method}\label{sec:4}
\subsection{Overview}\label{sec:41}
We present the networks to recurrently estimate semantic correspondences and synthesize the stylized images in a  boosting manner, as shown in~\figref{img:2}(c).  In the networks,  correspondences are robustly established by matching the stylized source and target images, in contrast to existing methods~\cite{Rocco18,Kim18nips} that directly match source and target images that have the attribute discrepancy. At the same time, blended neural patches using the correspondences are used to reconstruct the attribute transferred image in a semantic-aware and geometrically aligned manner. 

Our networks are split into three parts as shown in~\figref{img:3}: \emph{feature extraction networks} to extract source and target features $F^s$ and $F^t$, \emph{semantic matching network}s to establish correspondence fields $T$, and \emph{attribute transfer networks} to synthesize the attribute transferred image $I^{s\leftarrow t}$. Since our networks are formulated in a recurrent manner, they output $T^l$ and $I^{s\leftarrow t,l}$ at each $l$-th iteration, as exemplified in~\figref{img:6}.

\subsection{Network Architecture}\label{sec:42}
\paragraph{Feature extraction networks.}
Our model accomplishes the semantic matching and attribute transfer using deep features~\cite{Simonyan15,Kim18}.  
To extract the features for source $F^s$ and target $F^t$, the source and target images ($I^s$ and $I^t$) are first passed through shared feature extraction networks with parameters ${W}_F$ such that $F_i = \mathcal{F}(I_i;{W}_F)$, respectively. 
In the recurrent formulation, an attribute transferred feature $F^{s\leftarrow t,l}$ from target to source images and a warped target feature $F^{t,l}$, i.e., $F^{t}$ warped using the transformation fields ${T}^l_i$, are reconstructed at each $l$-th iteration. \vspace{-10pt}

\paragraph{Semantic matching networks.}
Our semantic matching networks consist of the matching cost computation and inference modules motivated by conventional RANSAC-like methods~\cite{Philbin07}. 
We first compute the correlation volume with respect to translational motion only~\cite{Rocco17,Rocco18,Rocco18nips,Kim18nips} and then pass it to subsequent convolutional layers to determine dense affine transformation fields $T_i$. 
%As shown in~\cite{Rocco18,Kim18nips}, this two-step approach reliably prunes incorrect matches. 

Unlike existing methods~\cite{Rocco17,Rocco18,Kim18nips}, our method computes the matching similarity between not only source and target features but also synthesized source and target features to minimize errors from the attribute discrepancy between source and target features such that:
\begin{equation}
\begin{split}
C^{l}_{i}(p) = 
&(1-\lambda^l)({F^s_i\boldsymbol{\cdot} F^{t,l}_p})/\|F^s_i\|\|F^{t,l}_p\|\\
&+\lambda^l({F^{s\leftarrow t,l}_i\boldsymbol{\cdot} F^{t,l}_p})/\|F^{s\leftarrow t,l}_i\|\|F^{t,l}_p\|,
\end{split}
\end{equation}
where ${p}\in P_i$ for local search window $P_i$ centered at $i$. $\lambda^l$ controls the trade-off between
content and attribute when computing the similarity, which is similar to~\cite{Liao17}. Note that when $\lambda^l=0$, we only consider the source feature $F^s$ without considering the stylized feature $F^{s\leftarrow t}$. These similarities undergo $L_2$ normalization to reduce errors~\cite{Rocco18}. 
\begin{figure}[t]
	\centering
	\renewcommand{\thesubfigure}{}
	\subfigure[(a)]
	{\includegraphics[width=0.33\linewidth]{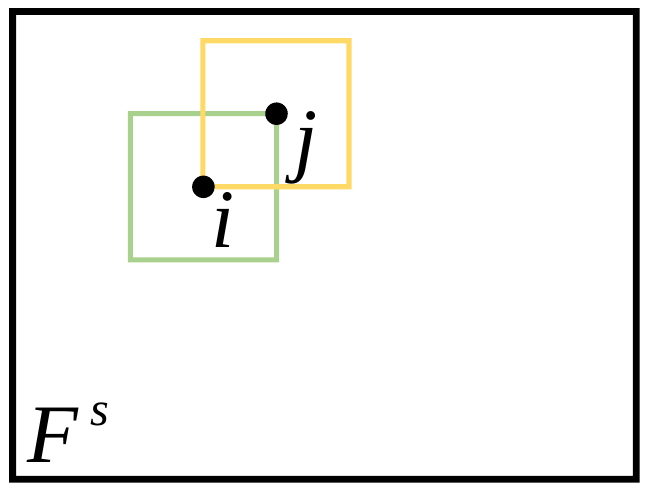}}\hfill
	\subfigure[(b)]
	{\includegraphics[width=0.33\linewidth]{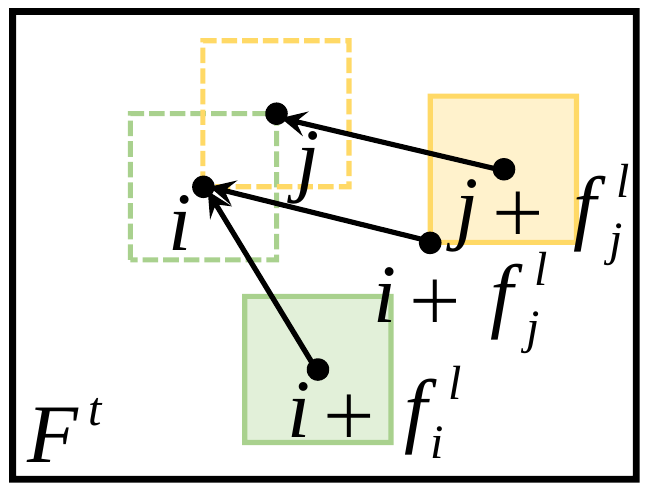}}\hfill
	\subfigure[(c)]
	{\includegraphics[width=0.33\linewidth]{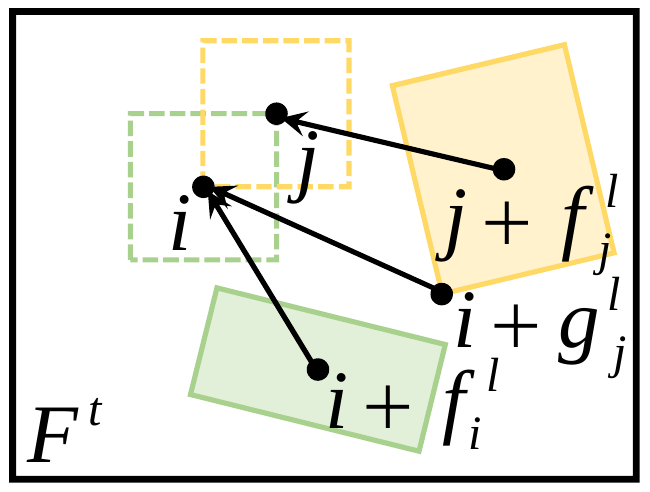}}\hfill
	\caption{Visualization of neural patch blending: for source feature $F^s$ in (a), unlike existing methods~\cite{Liao17,Li16,Gu18} that blend features of source $F^s$ and target $F^t$ using only traslationional fields $f_i$ as in (b), our method blends the features with the learned affine transformation fields $T^l_i = [{A}^l_{i},{f}^l_i]$ as in (c).}\label{img:4}\vspace{-10pt}
\end{figure} 

Based on this, the matching inference networks with parameters ${W}_G$ iteratively estimate the residual between the previous and current transformation fields~\cite{Kim18nips} as
\begin{equation}
{T}^{l}_i-{T}^{l-1}_i = \mathcal{F}(C^{l}_{i};{W}_G).
\end{equation}
The current transformation fields are then estimated in a recurrent manner~\cite{Kim18nips} as follows:
\begin{equation}
{T}^l_i = [\mathrm{I}_{2\times2},\mathrm{0}_{2\times1}] + \sum\nolimits_{n\in \phi(l)} {\mathcal{F}(C^{n}_{i};{W}_G)},
\end{equation} 
where $\phi(l) = \{1,..,{l-1}\}$. Unlike~\cite{Rocco17,Rocco18} that estimate a global affine or thin-plate spline transformation field, our networks are formulated as the encoder-decoder networks as in~\cite{Ronneberger15} to estimate locally-varying transformation fields. \vspace{-10pt}

\paragraph{Attribute transfer networks.}
To transfer the attribute of target feature $F^t$ into the content of source feature $F^s$ at $l$-th iteration, our attribute transfer networks first blend the source and target features as $B^{s\leftarrow t,l}$ using estimated transformation field ${T}^{l}_i$ and then reconstruct the stylized source image $I^{s\leftarrow t,l}$ using the decoder networks with parameters ${W}_D$ such that $I^{s\leftarrow t,l} = \mathcal{F}(B^{s\leftarrow t,l};{W}_D)$.
%, which will be used to extract the stylized features at next iteration such that $F^{s\leftarrow t,l+1} = \mathcal{F}(I^{s\leftarrow t,l};{W}_F)$. 

\begin{figure}[t]
	\centering
	\renewcommand{\thesubfigure}{}
	\subfigure[(a)]
	{\includegraphics[width=0.245\linewidth]{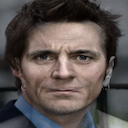}}\hfill
	\subfigure[(b)]
	{\includegraphics[width=0.245\linewidth]{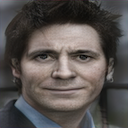}}\hfill
	\subfigure[(c)]
	{\includegraphics[width=0.245\linewidth]{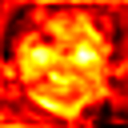}}\hfill
	\subfigure[(d)]
	{\includegraphics[width=0.245\linewidth]{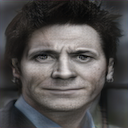}}\hfill
	\caption{Effects on the confidence in neural patch blending: (a) blending results of $I^s$ and $I^t$, (b) blending results of $F^s$ and $F^t$ followed by the decoder, (c) confidence, and (d) blending results of $F^s$ and $F^t$ with the confidence followed by the decoder.}\label{img:5}\vspace{-10pt}
\end{figure}

Specifically, our neural patch blending between $F^s$ and $F^t$ with the current transformation field $T^l_i = [{A}^l_{i},{f}^l_i]$ is formulated as shown in~\figref{img:4} such that
\begin{equation}
B^{s\leftarrow t,l}_i = (1-\lambda^l)F^s_i+\lambda^l\sum\nolimits_{j\in N_i}{\alpha^l_j F^{t}_{i+g^l_j}}/\sum\nolimits_{j\in N_i}{\alpha^l_j},
\end{equation}
where $g^l_j = (A^l_j-\mathrm{I}_{2\times2})(i-j)+f^l_j$. $\alpha^l_i$ is a confidence of each pixel $i$ that has $T^l_i$ computed similar to~\cite{Kim18conf} such that
\begin{equation}
\alpha^l_i = \exp(C^l_{i}(i))/\sum\nolimits_{p\in P_i}{\exp(C^l_{i}(p))}.
\end{equation}
Our neural patch blending module differs from the existing methods~\cite{Liao17,Li16,Gu18} in the use of learned transformation fields and consideration of more complex deformations such as affine transformations. In addition, unlike exisiting style transfer methods~\cite{Li16,Gu18}, our networks employ the confidence to transfer the attribute of matchable points only tailored to our objective, as exemplified in~\figref{img:5}.

In addition, our decoder networks are formulated as a symmetric structure to feature extraction networks. 
%Unlike existing greedy optimization based methods~\cite{Gatys15,Li16,Luan17}, the feed-forward decoder networks synthesize an image very efficiently and enable an end-to-end training~\cite{Chen16,Johnson16}. 
Since the single-level decoder networks as in~\cite{Huang17} cannot capture both complicated structures at high-level features and low-level information at low-level features, the multi-level decoder networks have been proposed as in~\cite{Li17,Li18}, but they are not very economic~\cite{Gu18}. 
Instead, we use the skip connection from the source features $F^s$ to capture both low- and high-level attribute characteristics~\cite{Li17,Li18,Gu18}. However, using the skip connection through simple concatenation~\cite{Ronneberger15} makes the decoder networks reconstruct an image using only low-level features. To alleviate this, inspired by a dropout layer~\cite{Srivastava14}, we present a droplink layer such that the skipped features and upsampled features are stochastically linked to avoid the overfitting to certain level features:
\begin{equation}
F^{s\leftarrow t,l}_h = 
(1-b_h)\mathcal{F}(B^{s\leftarrow t,l};{W}_{D,h})+b_h F^s_h,
\end{equation}
where $F^{s\leftarrow t,l}_h$ and $F^s_h$ are the intermediate and skipped features at $h$-th level for $h\in \{1,...,H\}$. ${W}_{D,h}$ is the parameters until $h$-th level. $b_h$ is a binary random variable. Note that if $b_h=0$, this becomes the no-skip connected layer.  

\subsection{Loss Functions}\label{sec:43}
\paragraph{Semantic attribute matching loss.}
Our networks are learned using weak supervision in the form of image pairs. Concretely, we present a semantic attribute matching loss in a manner that the transformation field $T$ and the stylized image $I^{s\leftarrow t}$ can be simultaneously learned and inferred to minimize a single loss function. After the convergence of iterations at $L$-th iteration, an attribute transferred feature $F^{s\leftarrow t,L}$ and a warped target feature $F^{t,L}$ are used to define the loss function. This intuition can be realized by minimizing the following objective:
\begin{equation}\label{equ:sim_loss}
D(F^{s\leftarrow t,L},F^{t,L}) = \sum\limits_{i}{\sum\limits_{j\in N_i}{\|F^{s\leftarrow t,L}_j-F^{t,L}_j\|^2_F}}.
\end{equation}
In comparison to existing the matching loss $\mathcal{L}_{M}$ and the attribute transfer loss $\mathcal{L}_{A}$, this objective enables us to solve the photometric and geometric variations across semantically similar images simultaneously.

Although using only this objective provides satisfactory performance, we extend this objective to consider both positive and negative samples to enhance network training and precise localization ability based on the intuition that the matching score should be minimized at the correct transformation while keeping the scores of other neighbor transformation candidates high. Finally, we formulate our semantic attribute matching loss as a cross-entropy loss as
\begin{equation}
\mathcal{L}_{AM}=
\sum\nolimits_{i}{\max\left(-\log(K_{i}),\tau\right)},
\end{equation}
where $K_{i}$ is the softmax probability defined as
\begin{equation}\label{equ:prob_corr}
K_{i} = \frac{\exp(-D(F^{s\leftarrow t,L}_i,F^{t,L}_i))}
{{\sum\nolimits_{q\in Q_i} {\exp (-D(F^{s\leftarrow t,L}_i,F^{t,L}_q))} }}.
\end{equation}
It makes the center point $i$ within the neighbor $Q_i$ become a positive sample and the other points become negative samples. In addition, the truncated max operator $\max(\cdot,\tau)$ is used to focus on the sailent parts such as objects during training with the parameter $\tau$. \vspace{-10pt}
\begin{figure}[t]
	\centering
	\renewcommand{\thesubfigure}{}
	\subfigure[]
	{\includegraphics[width=0.7\linewidth]{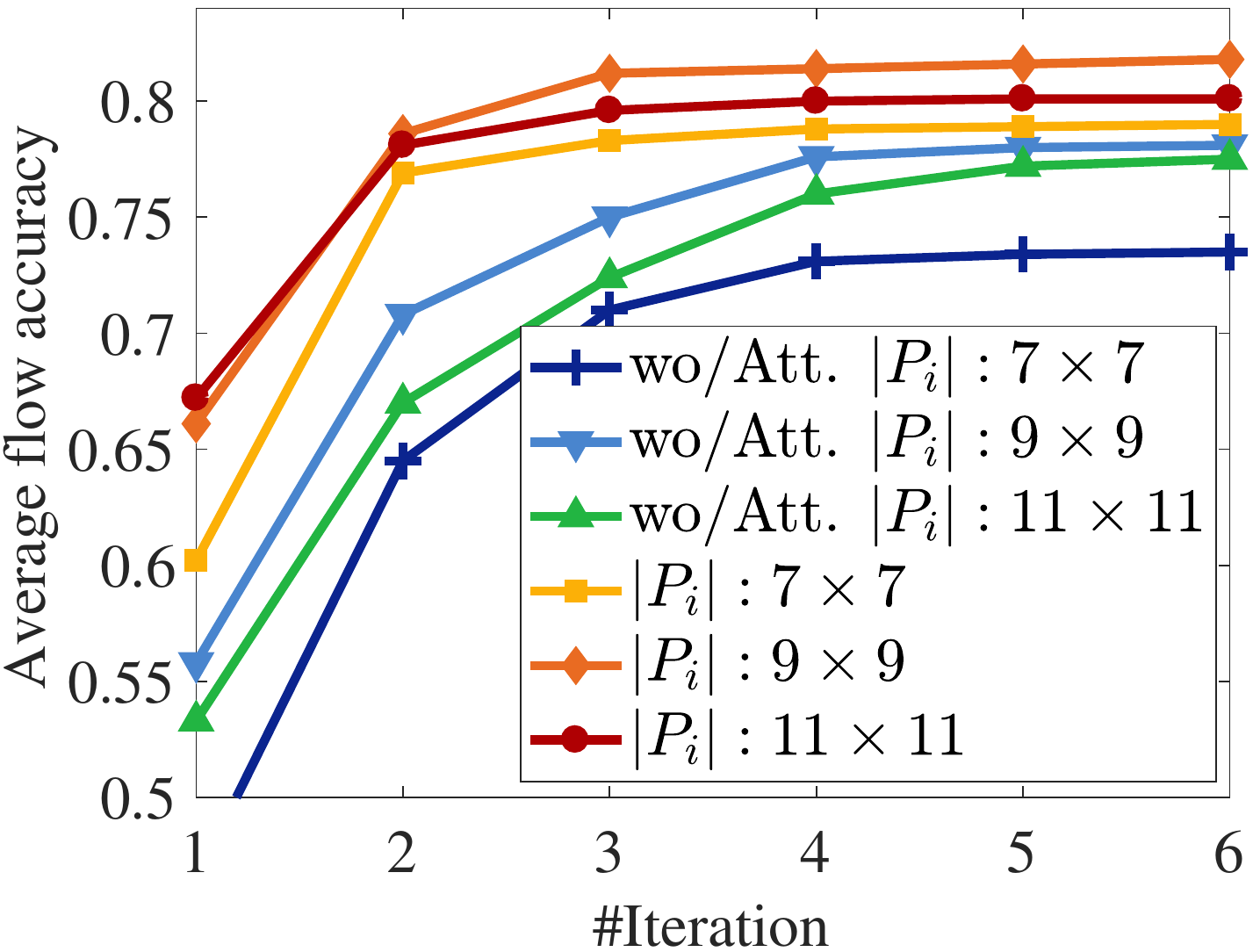}}\\
	\vspace{-10pt}
	\caption{Convergence analysis of SAM-Net for various numbers of iterations and search window sizes on the TSS benchmark~\cite{Taniai16}.}\label{img:7}\vspace{-10pt}
\end{figure}
\begin{figure}[t]
	\centering
	\renewcommand{\thesubfigure}{}
	\subfigure[]
	{\includegraphics[width=0.245\linewidth]{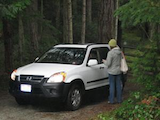}}\hfill
	\subfigure[]
	{\includegraphics[width=0.245\linewidth]{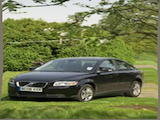}}\hfill
	\subfigure[]
	{\includegraphics[width=0.245\linewidth]{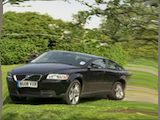}}\hfill
	\subfigure[]
	{\includegraphics[width=0.245\linewidth]{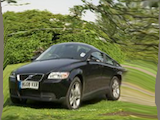}}\hfill\\
	\vspace{-21.5pt}
	\subfigure[(a) input images]
	{\includegraphics[width=0.245\linewidth]{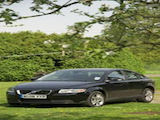}}\hfill
	\subfigure[(b) iter 1]
	{\includegraphics[width=0.245\linewidth]{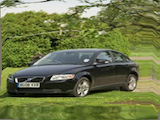}}\hfill
	\subfigure[(c) iter 2]
	{\includegraphics[width=0.245\linewidth]{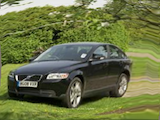}}\hfill
	\subfigure[(d) iter 3]
	{\includegraphics[width=0.245\linewidth]{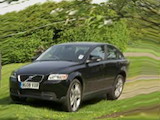}}\hfill\\
	\caption{Ablation study of SAM-Net without (top) and with (bottom) attribute transfer networks as evolving iterations.}\label{img:8}\vspace{-10pt}
\end{figure}

\paragraph{Other losses.}
We utilize two additional losses, namely the content loss $\mathcal{L}_{C}$ as in~\equref{equ:loss_cont} to preserve the structure of source image and the $L_2$ regularization loss~\cite{Johnson16,Li16} to encourage spatial smoothness in the stylized image.

\section{Experiments}\label{sec:5}
\subsection{Training and Implementation Details}\label{sec:44}
To learn our SAM-Net, large-scale semantically similar image pairs are needed, but such public datasets are limited quantitatively. To overcome this, we adopt a two-step training technique, similar to~\cite{Rocco18}. In the first step, we train our networks using a synthetic training dataset provided in~\cite{Rocco17}, where synthetic transformations are randomly applied to a single image to generate the image pairs, and thus the images do not have appearance variations. 
%Existing some style transfer methods~\cite{Li17,Huang17,Li18} also independently train their decoder networks similar to our first-step. 
This enables the attribute transfer networks to be learned in an auto-encoder manner~\cite{Li17,Huang17,Li18}, but the matching networks still have limited ability to deal with the attribute variations. To overcome this, in the second step, we finetune this pretrained network on public datasets for semantically similar image pairs from the training set of PF-PASCAL~\cite{Ham17} following the split used in~\cite{Ham17}. 
\begin{table}[t]
	\begin{center}
		\begin{tabular}{ >{\raggedright}m{0.34\linewidth}
				>{\centering}m{0.10\linewidth} >{\centering}m{0.10\linewidth}
				>{\centering}m{0.10\linewidth} >{\centering}m{0.10\linewidth}}
			\hlinewd{0.8pt}
			Methods &FG3D &JODS &PASC. &Avg.\tabularnewline
			\hline
			\hline
			Taniai et al.~\cite{Taniai16} &0.830 &0.595 &0.483 &0.636\tabularnewline
			PF~\cite{Ham16} &0.786 &0.653 &0.531 &0.657 \tabularnewline
			%UCN~\cite{Choy16} &0.853 &0.672 &0.511 &0.679 \tabularnewline
			%FCSS~\cite{Kim17} &0.858 &0.680 &0.522 &0.687 \tabularnewline
			DCTM~\cite{Kim17c} &0.891 &0.721 &0.610 &0.740 \tabularnewline
			SCNet~\cite{Han17} &0.776 &0.608 &0.474 &0.619 \tabularnewline
			GMat.~\cite{Rocco17} &0.835 &0.656 &0.527 &0.673 \tabularnewline
			GMat. w/Inl.~\cite{Rocco18} &0.892 &0.758 &0.562 &0.737 \tabularnewline
			DIA~\cite{Liao17} &0.762 &0.685 &0.513 &0.653 \tabularnewline
			RTNs~\cite{Kim18nips} &0.901 &0.782 &0.633 &0.772 \tabularnewline
			\hline
			SAM-Net w/\equref{equ:sim_loss} &0.891 &0.789 &0.638 &0.773 \tabularnewline
			SAM-Net wo/Att. &0.912 &0.790 &0.641 &0.781 \tabularnewline
			SAM-Net &\bf{0.961} &\bf{0.822} &\bf{0.672} &\bf{0.818} \tabularnewline
			\hlinewd{0.8pt}
		\end{tabular}
	\end{center}
	\vspace{-5pt}
	\caption{Matching accuracy compared to the state-of-the-art correspondence techniques on the TSS benchmark~\cite{Taniai16}.}\label{tab:1}\vspace{-5pt}
\end{table}
\begin{table}[!t]
	\begin{center}
		\begin{tabular}{ >{\raggedright}m{0.31\linewidth}
				>{\centering}m{0.16\linewidth} >{\centering}m{0.16\linewidth}
				>{\centering}m{0.16\linewidth}}
			\hlinewd{0.8pt}
			\multirow{2}{*}{Methods}
			&\multicolumn{3}{ c }{PCK} \tabularnewline
			\cline{2-4}
			&$\alpha=0.05$ &$\alpha=0.1$ &$\alpha=0.15$\tabularnewline
			\hline
			\hline
			PF~\cite{Ham16} &0.314 &0.625 &0.795 \tabularnewline
			%UCN~\cite{Choy16} &0.299 &0.556 &0.740 \tabularnewline
			%FCSS~\cite{Kim18} &0.336 &0.689 &0.792 \tabularnewline
			DCTM~\cite{Kim17c} &0.342 &0.696 &0.802 \tabularnewline
			SCNet~\cite{Han17} &0.362 &0.722 &0.820 \tabularnewline	
			GMat.~\cite{Rocco17} &0.410 &0.695 &0.804 \tabularnewline
			GMat. w/Inl.~\cite{Rocco18} &0.490 &0.748 &0.840 \tabularnewline
			DIA~\cite{Liao17} &0.471 &0.724 &0.811 \tabularnewline
			RTNs~\cite{Kim18nips} &0.552 &0.759 &0.852 \tabularnewline
			NC-Net~\cite{Rocco18nips} &- &0.789 &- \tabularnewline
			\hline
			SAM-Net &\bf{0.601} &\bf{0.802} &\bf{0.869} \tabularnewline
			\hlinewd{0.8pt}
		\end{tabular}
	\end{center}
	\vspace{-5pt}
	\caption{Matching accuracy compared to the state-of-the-art correspondence techniques on the PF-PASCAL benchmark~\cite{Ham17}.}\label{tab:2}\vspace{-10pt}
\end{table}
\begin{figure*}[t]
	\centering
	\renewcommand{\thesubfigure}{}
	\subfigure[]
	{\includegraphics[width=0.122\linewidth]{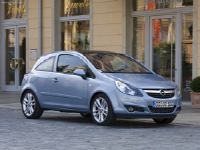}}\hfill
	\subfigure[]
	{\includegraphics[width=0.122\linewidth]{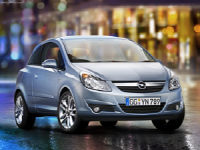}}\hfill
	\subfigure[]
	{\includegraphics[width=0.122\linewidth]{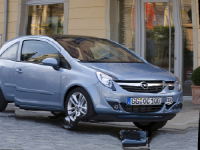}}\hfill
	\subfigure[]
	{\includegraphics[width=0.122\linewidth]{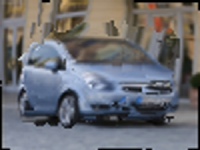}}\hfill
	\subfigure[]
	{\includegraphics[width=0.122\linewidth]{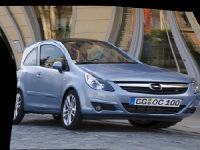}}\hfill
	\subfigure[]
	{\includegraphics[width=0.122\linewidth]{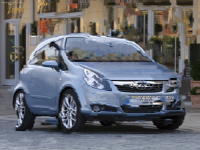}}\hfill
	\subfigure[]
	{\includegraphics[width=0.122\linewidth]{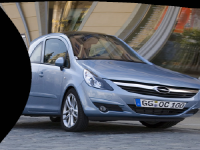}}\hfill
	\subfigure[]
	{\includegraphics[width=0.122\linewidth]{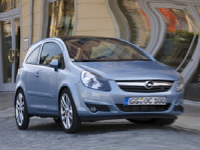}}\hfill\\
	\vspace{-21.5pt}
	\subfigure[(a)]
	{\includegraphics[width=0.122\linewidth]{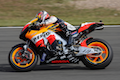}}\hfill
	\subfigure[(b)]
	{\includegraphics[width=0.122\linewidth]{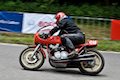}}\hfill
	\subfigure[(c)]
	{\includegraphics[width=0.122\linewidth]{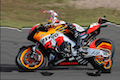}}\hfill
	\subfigure[(d)]
	{\includegraphics[width=0.122\linewidth]{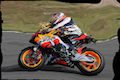}}\hfill
	\subfigure[(e)]
	{\includegraphics[width=0.122\linewidth]{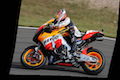}}\hfill
	\subfigure[(f)]
	{\includegraphics[width=0.122\linewidth]{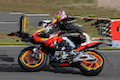}}\hfill
	\subfigure[(g)]
	{\includegraphics[width=0.122\linewidth]{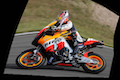}}\hfill
	\subfigure[(h)]
	{\includegraphics[width=0.122\linewidth]{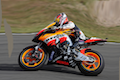}}\hfill\\
	\caption{Qualitative results on the TSS benchmark~\cite{Taniai16}: (a) source and (b) target images, warped source images using correspondences of 
		(c) PF~\cite{Ham16}, 
		(d) DCTM~\cite{Kim17c}, 
		(e) GMat~\cite{Rocco17}, 
		(f) DIA~\cite{Liao17}, (g) GMat. w/Inl.~\cite{Rocco18}, and (h) SAM-Net.}\label{img:9}\vspace{-10pt}
\end{figure*}
\begin{figure*}[t]
	\centering
	\renewcommand{\thesubfigure}{}
	\subfigure[]
	{\includegraphics[width=0.122\linewidth]{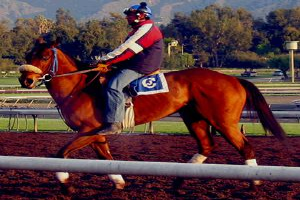}}\hfill
	\subfigure[]
	{\includegraphics[width=0.122\linewidth]{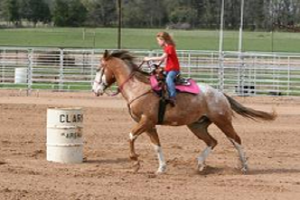}}\hfill
	\subfigure[]
	{\includegraphics[width=0.122\linewidth]{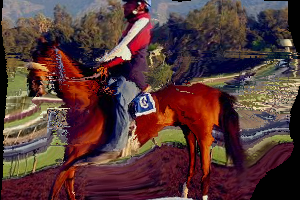}}\hfill
	\subfigure[]
	{\includegraphics[width=0.122\linewidth]{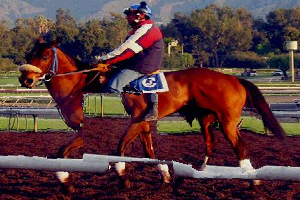}}\hfill
	\subfigure[]
	{\includegraphics[width=0.122\linewidth]{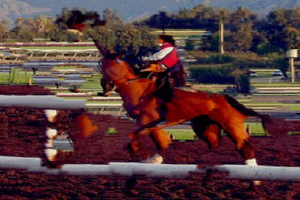}}\hfill
	\subfigure[]
	{\includegraphics[width=0.122\linewidth]{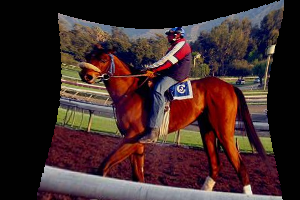}}\hfill
	\subfigure[]
	{\includegraphics[width=0.122\linewidth]{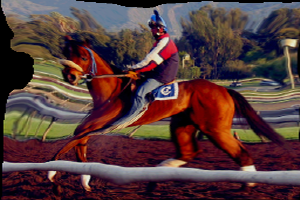}}\hfill
	\subfigure[]
	{\includegraphics[width=0.122\linewidth]{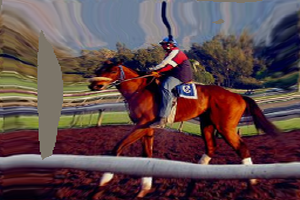}}\hfill\\
	\vspace{-21.5pt}
	\subfigure[(a)]
	{\includegraphics[width=0.122\linewidth]{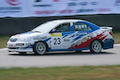}}\hfill
	\subfigure[(b)]
	{\includegraphics[width=0.122\linewidth]{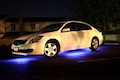}}\hfill
	\subfigure[(c)]
	{\includegraphics[width=0.122\linewidth]{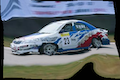}}\hfill
	\subfigure[(d)]
	{\includegraphics[width=0.122\linewidth]{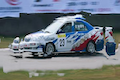}}\hfill
	\subfigure[(e)]
	{\includegraphics[width=0.122\linewidth]{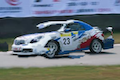}}\hfill
	\subfigure[(f)]
	{\includegraphics[width=0.122\linewidth]{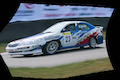}}\hfill
	\subfigure[(g)]
	{\includegraphics[width=0.122\linewidth]{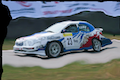}}\hfill
	\subfigure[(h)]
	{\includegraphics[width=0.122\linewidth]{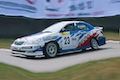}}\hfill\\
	\caption{Qualitative results on the PF-PASCAL benchmark~\cite{Ham16}: (a) source and (b) target images, warped source images using correspondences of (c) DCTM~\cite{Kim17c}, (d) SCNet~\cite{Han17}, (e) DIA~\cite{Liao17} (f) GMat. w/Inl.~\cite{Rocco18}, (g) RTNs~\cite{Kim18nips}, and (h) SAM-Net.}\label{img:10}\vspace{-10pt}
\end{figure*}

For feature extraction, we used the ImageNet-pretrained VGG-19 networks~\cite{Simonyan15}, where the activations are extracted from `relu4-1' layer (i.e., $H=4$). We gradually increase $\lambda^l$ until 1 such that $\lambda^l=1-\exp(-l)$. During training, we set the maximum number of iteration $L$ to 5 to avoid the gradient vanishing and exploding problem. During testing, the iteration count is increased to 10. Following~\cite{Kim18nips}, the window sizes of $N_i$, $P_i$, and $Q_i$ are set to $3\times 3$, $9\times 9$, and $9\times 9$, respectively. The probability of $b_h$ is defined as 0.9 and in testing $b_h$ is set to 0.5. 

\subsection{Experimental Settings}\label{sec:51}
In the following, we comprehensively evaluated SAM-Net through comparisons to state-of-the-art methods for semantic matching,
including Taniai et al.~\cite{Taniai16}, PF~\cite{Ham16}, SCNet~\cite{Han17}, DCTM~\cite{Kim17}, DIA~\cite{Liao17}, GMat.~\cite{Rocco17}, GMat. w/Inl.~\cite{Rocco18}, NC-Net~\cite{Rocco18nips}, RTNs~\cite{Kim18nips}, and for attribute transfer, including Gatys et al.~\cite{Gatys15}, CNN-MRF~\cite{Li16}, Photo-WCT~\cite{Li18}, Gu et al.~\cite{Gu18}, and DIA~\cite{Liao17}. Performance was measured on TSS dataset~\cite{Taniai16}, PF-PASCAL dataset~\cite{Ham17}, and CUB-200-2011 dataset~\cite{WahCUB_200_2011}. 

In~\secref{sec:52}, we first analyzed the effects of the components within SAM-Net, and then evaluated matching results with various benchmarks and quantitative measures in~\secref{sec:53}. We finally evaluated photorealistic attribute transfer results with various applications in~\secref{sec:54}.

\subsection{Ablation Study}\label{sec:52}
To validate the components within SAM-Net, we evaluated the matching accuracy for different numbers of iterations, with various sizes of $P_i$, and with and without attribute transfer module. For quantitative assessment, we examined the accuracy on the TSS benchmark~\cite{Taniai16}. As shown in~\figref{img:7},~\figref{img:8}, and \tabref{tab:1}, SAM-Net converges  in 2$-$3 iterations. In addition, the results of `SAM-Net wo/Att.', i.e., SAM-Net without attribute transfer, show the effectiveness of attribute transfer module in the recurrent formulation. The results of `SAM-Net wo/\equref{equ:sim_loss}.', i.e., SAM-Net with the loss of \equref{equ:sim_loss}, show the importance to consider the negative samples when training. By enlarging the size of $P_i$, the accuracy improves until 9$\times$9, but larger window sizes reduce matching accuracy due to greater matching ambiguity. Note that $Q_i=P_i$ following to~\cite{Kim18nips}.

\subsection{Semantic Matching Results}\label{sec:53}
\paragraph{TSS benchmark.}
We evaluated SAM-Net on the TSS benchmark~\cite{Taniai16}, consisting of 400 image pairs.
As in~\cite{Kim17,Kim17c}, flow accuracy was
measured in~\tabref{tab:1}. \figref{img:9} shows qualitative results. Unlike existing methods~\cite{Choy16,Taniai16,Ham16,Han17,Kim17,Rocco17,Rocco18,Kim18nips} that do not consider the attribute variations between semantically similar images, our SAM-Net has shown highly improved preformance qualitatively and quantitatively. DIA~\cite{Liao17} has shown limited matching accuracy compared to other deep methods~\cite{Rocco18,Kim18nips}, due to their limited regularization powers. 
Unlike this, the results of our SAM-Net shows that our method is more successfully transferring the attribute between source and target images to improve the semantic matching accuracy. 
\vspace{-10pt}
\begin{figure*}[t]
	\centering
	\renewcommand{\thesubfigure}{}
	\subfigure[]
	{\includegraphics[width=0.122\linewidth]{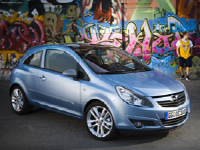}}\hfill
	\subfigure[]
	{\includegraphics[width=0.122\linewidth]{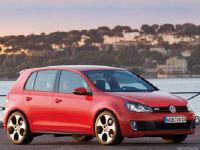}}\hfill
	\subfigure[]
	{\includegraphics[width=0.122\linewidth]{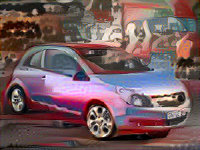}}\hfill
	\subfigure[]
	{\includegraphics[width=0.122\linewidth]{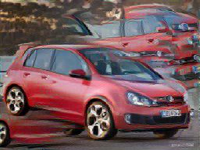}}\hfill
	\subfigure[]
	{\includegraphics[width=0.122\linewidth]{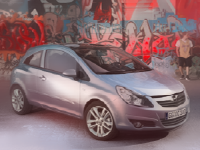}}\hfill
	\subfigure[]
	{\includegraphics[width=0.122\linewidth]{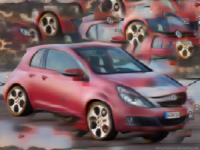}}\hfill
	\subfigure[]
	{\includegraphics[width=0.122\linewidth]{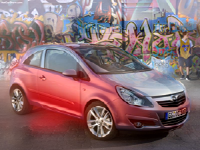}}\hfill
	\subfigure[]
	{\includegraphics[width=0.122\linewidth]{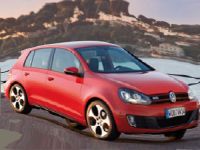}}\hfill\\
	\vspace{-21.5pt}
	\subfigure[]
	{\includegraphics[width=0.122\linewidth]{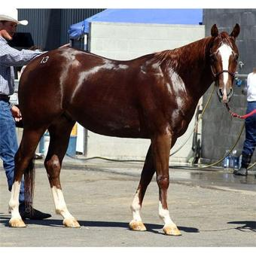}}\hfill
	\subfigure[]
	{\includegraphics[width=0.122\linewidth]{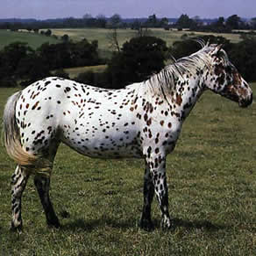}}\hfill
	\subfigure[]
	{\includegraphics[width=0.122\linewidth]{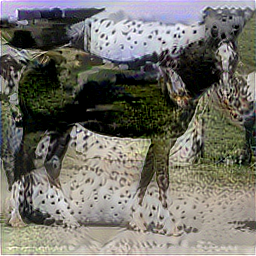}}\hfill
	\subfigure[]
	{\includegraphics[width=0.122\linewidth]{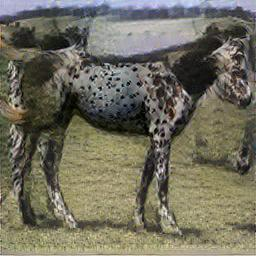}}\hfill
	\subfigure[]
	{\includegraphics[width=0.122\linewidth]{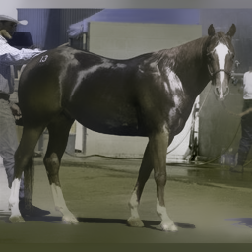}}\hfill
	\subfigure[]
	{\includegraphics[width=0.122\linewidth]{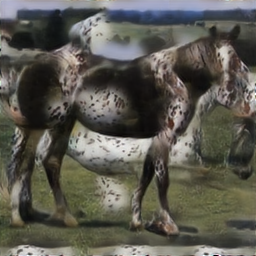}}\hfill
	\subfigure[]
	{\includegraphics[width=0.122\linewidth]{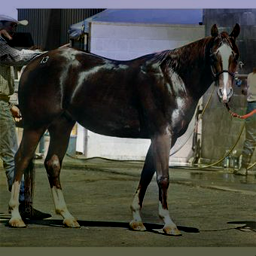}}\hfill
	\subfigure[]
	{\includegraphics[width=0.122\linewidth]{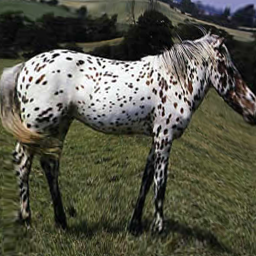}}\hfill\\
	\vspace{-21.5pt}
	\subfigure[]
	{\includegraphics[width=0.122\linewidth]{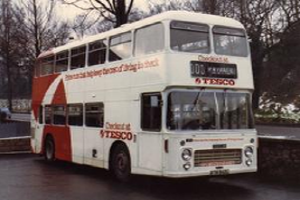}}\hfill
	\subfigure[]
	{\includegraphics[width=0.122\linewidth]{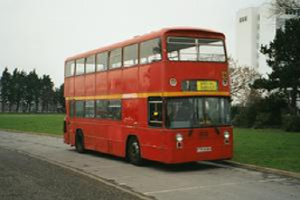}}\hfill
	\subfigure[]
	{\includegraphics[width=0.122\linewidth]{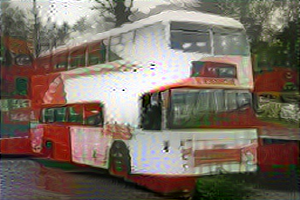}}\hfill
	\subfigure[]
	{\includegraphics[width=0.122\linewidth]{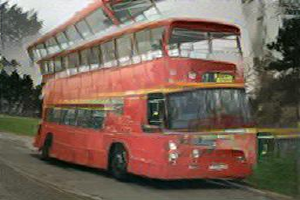}}\hfill
	\subfigure[]
	{\includegraphics[width=0.122\linewidth]{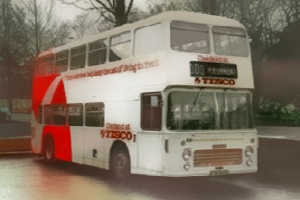}}\hfill
	\subfigure[]
	{\includegraphics[width=0.122\linewidth]{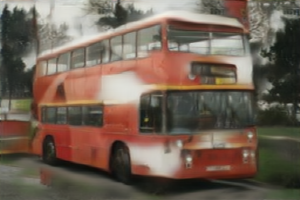}}\hfill
	\subfigure[]
	{\includegraphics[width=0.122\linewidth]{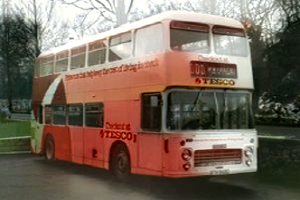}}\hfill
	\subfigure[]
	{\includegraphics[width=0.122\linewidth]{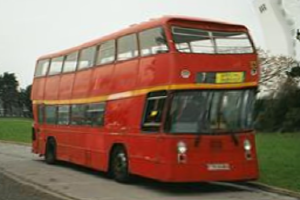}}\hfill\\
	\vspace{-21.5pt}
	\subfigure[(a)]
	{\includegraphics[width=0.122\linewidth]{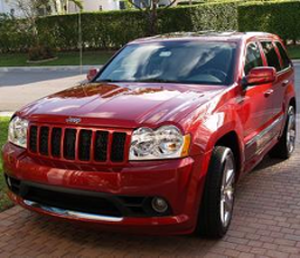}}\hfill
	\subfigure[(b)]
	{\includegraphics[width=0.122\linewidth]{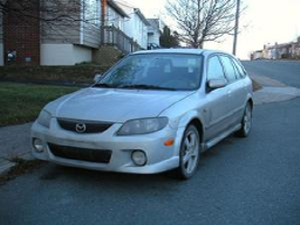}}\hfill
	\subfigure[(c)]
	{\includegraphics[width=0.122\linewidth]{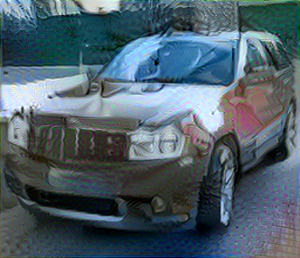}}\hfill
	\subfigure[(d)]
	{\includegraphics[width=0.122\linewidth]{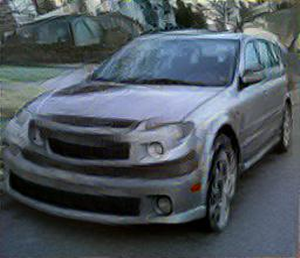}}\hfill
	\subfigure[(e)]
	{\includegraphics[width=0.122\linewidth]{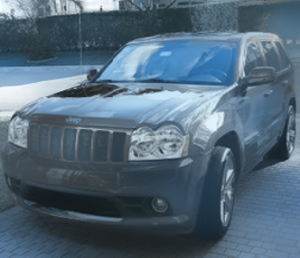}}\hfill
	\subfigure[(f)]
	{\includegraphics[width=0.122\linewidth]{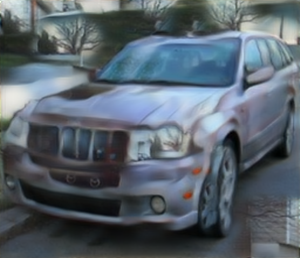}}\hfill
	\subfigure[(g)]
	{\includegraphics[width=0.122\linewidth]{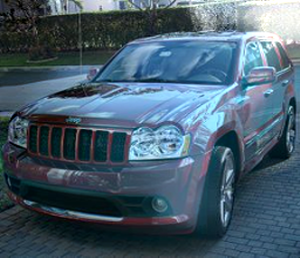}}\hfill
	\subfigure[(h)]
	{\includegraphics[width=0.122\linewidth]{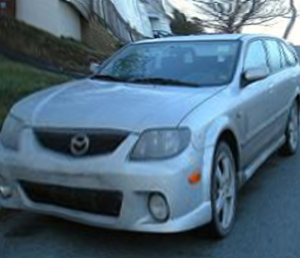}}\hfill\\
	\caption{Qualitative results of the photorealistic attribute transfer on the TSS~\cite{Taniai16} PF-PASCAL~\cite{Ham17} benchmarks: (a) source and (b) target images, results of (c) Gatys et al.~\cite{Gatys15}, (d) CNN-MRF~\cite{Li16}, (e) Photo-WCT~\cite{Li18}, (f) Gu et al.~\cite{Gu18}, (g) DIA~\cite{Liao17}, and (h) SAM-Net.}\label{img:11}\vspace{-10pt}
\end{figure*}

\paragraph{PF-PASCAL benchmark.}
We also evaluated SAM-Net on the PF-PASCAL benchmark~\cite{Ham17},
which contains 1,351 image pairs over 20 object categories with PASCAL keypoint annotations~\cite{Bourdev09}. %Following the split in~\cite{Han17,Rocco18}, we used 700 training pairs, 300 validation pairs, and 300 testing pairs. 
%Note that this benchmark has more severe photometric and geometric variations in comparison to the TSS benchmark~\cite{Taniai16}. 
For the evaluation metric, we used the PCK between flow-warped keypoints and the ground truth as done in the experiments of~\cite{Han17}. 
\tabref{tab:2} summarizes the PCK values, and \figref{img:10} shows qualitative results. Similar to the experiments on the TSS benchmark~\cite{Taniai16}, CNN-based methods~\cite{Han17,Rocco17,Rocco18,Rocco18,Kim18nips} including our SAM-Net
yield better performance, with SAM-Net providing the highest matching accuracy.
\begin{figure}[t]
	\centering
	\renewcommand{\thesubfigure}{}
	\subfigure[]
	{\includegraphics[width=0.195\linewidth]{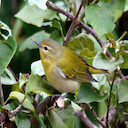}}\hfill
	\subfigure[]
	{\includegraphics[width=0.195\linewidth]{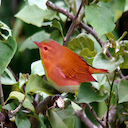}}\hfill
	\subfigure[]
	{\includegraphics[width=0.195\linewidth]{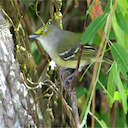}}\hfill
	\subfigure[]
	{\includegraphics[width=0.195\linewidth]{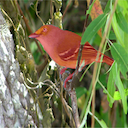}}\hfill
	\subfigure[]
	{\includegraphics[width=0.195\linewidth]{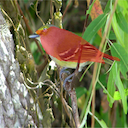}}\hfill\\
	\vspace{-21.5pt}
	\subfigure[(a)]
	{\includegraphics[width=0.195\linewidth]{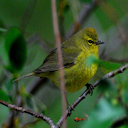}}\hfill
	\subfigure[(b)]
	{\includegraphics[width=0.195\linewidth]{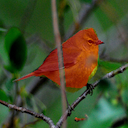}}\hfill
	\subfigure[(c)]
	{\includegraphics[width=0.195\linewidth]{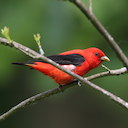}}\hfill
	\subfigure[(d)]
	{\includegraphics[width=0.195\linewidth]{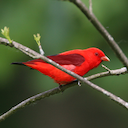}}\hfill
	\subfigure[(e)]
	{\includegraphics[width=0.195\linewidth]{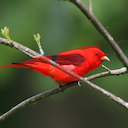}}\hfill\\
	\caption{Qualitative results of the mask transfer on the CUB-200-2011 benchmark~\cite{WahCUB_200_2011}: source (a) images and (b) masks and target (c) images and (d) masks, and (e) warped source masks to the target images using correspondences from SAM-Net.}\label{img:12}\vspace{-10pt}
\end{figure}

\subsection{Applications}\label{sec:54}
\paragraph{Photorealistic attribute transfer.}
We evaluated SAM-Net for photorealistic attribute transfer on the TSS~\cite{Taniai16} and PF-PASCAL benchmarks~\cite{Ham17}. For evaluatation, we sampled the image pairs from these datasets and transferred the attribute of target image to the source image as shown in~\figref{img:11}. Note that SAM-Net is designed to work on images contain that semantically similar contents and not effective for generic artistic style transfer applications as in~\cite{Gatys15, Johnson16, Huang17}. As expected, existing methods tailored to artistic stylization such as a method of Gatys et al.~\cite{Gatys15} and CNN-MRF~\cite{Li16} produce limited quality images. Moreover, recent photorealistic stylization methods such as Photo-WCT~\cite{Li18} and Gu et al.~\cite{Gu18} have limited performance for the images that have background clutters. DIA~\cite{Liao17} provided degraded results due to its weak regularization technique. Unlike these methods, our SAM-Net has shown highly accurate and plausible results thanks to their learned transformation fields to synthesize the images. Note that some methods such as Photo-WCT~\cite{Li18} and DIA~\cite{Liao17} have used to refine their results using additional smoothing modules, but SAM-Net does not use any post-processing. 
\vspace{-10pt}
\begin{figure}[t]
	\centering
	\renewcommand{\thesubfigure}{}
	\subfigure[]
	{\includegraphics[width=0.195\linewidth]{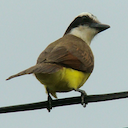}}\hfill
	\subfigure[]
	{\includegraphics[width=0.195\linewidth]{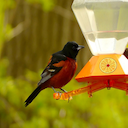}}\hfill
	\subfigure[]
	{\includegraphics[width=0.195\linewidth]{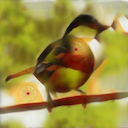}}\hfill
	\subfigure[]
	{\includegraphics[width=0.195\linewidth]{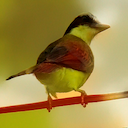}}\hfill
	\subfigure[]
	{\includegraphics[width=0.195\linewidth]{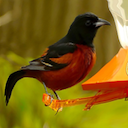}}\hfill\\
	\vspace{-21.5pt}
	\subfigure[(a)]
	{\includegraphics[width=0.195\linewidth]{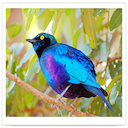}}\hfill
	\subfigure[(b)]
	{\includegraphics[width=0.195\linewidth]{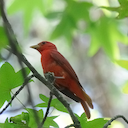}}\hfill
	\subfigure[(c)]
	{\includegraphics[width=0.195\linewidth]{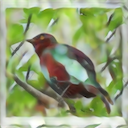}}\hfill
	\subfigure[(d)]
	{\includegraphics[width=0.195\linewidth]{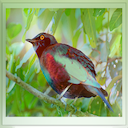}}\hfill
	\subfigure[(e)]
	{\includegraphics[width=0.195\linewidth]{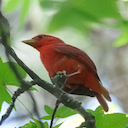}}\hfill\\
	\caption{Qualitative results of the object transfiguration on the CUB-200-2011 benchmark~\cite{WahCUB_200_2011}: (a) source and (b) target images, results of (c) Gu et al.~\cite{Gu18}, (d) DIA~\cite{Liao17}, and (e) SAM-Net. }\label{img:13}\vspace{-10pt}
\end{figure}

\paragraph{Foreground mask transfer.}
We evaluated SAM-Net for mask transfer on the CUB-200-2011 dataset~\cite{WahCUB_200_2011}, which contains images of 200 bird categories, with annotated foreground masks.
For semantically similar images that have very challenging photometric and geometric variations, our SAM-Net successfully transfers the semantic labels, as shown in~\figref{img:12}.
\vspace{-10pt}

\paragraph{Object transfiguration.}
We finally applied our method to object transfiguration, e.g., translating a source bird into a target breed. We used object classes from the CUB-200-2011 dataset~\cite{WahCUB_200_2011}. In this application, our SAM-Net has shown very plausible results as exemplified in~\figref{img:13}. 

\section{Conclusion}\label{sec:6}
We presented SAM-Net that recurrently estimates dense correspondences and transfers the attributes across semantically similar images in a joint and boosting manner. 
The key idea of this approach is to formulate the semantic matching and attribute transfer networks to complement each other through an iterative process.
For weakly-supervised training of SAM-Net, the semantic attribute matching loss is presented, which enables us to alleviate the photometric and geometric variations across the images simultaneously. 

{\small
	\bibliographystyle{ieee}
	\bibliography{egbib}
}

\end{document}